\documentclass[11pt,letterpaper]{article}
\usepackage{emnlp2017}
\usepackage{times}
\usepackage{latexsym}

\usepackage{color}
\usepackage{booktabs}
\usepackage{multirow}
\usepackage{amsmath}
\usepackage{amssymb}
\usepackage{url}
\usepackage{graphicx}
\usepackage{tipa}
\usepackage{subcaption}
\usepackage{enumitem}

\newcommand{\hl}[1]{\textbf{#1}}
\newcommand{\hll}[1]{\underline{#1}}
\newcommand{\hh}[1]{\textbf{#1}}

\newcommand{\eg}{\textit{e.g.\,}}
\newcommand{\vs}{\textit{vs.\,}}
\newcommand{\etc}{\textit{etc.\,}}
\newcommand{\viz}{\textit{viz.\,}}
\newcommand{\ie}{\textit{i.e.\,}}

\newcommand*{\thead}[1]{\multicolumn{1}{c}{\bfseries #1}}

\emnlpfinalcopy

\def\emnlppaperid{24}

\title{Learning variable length units for SMT between related languages via Byte Pair Encoding}

\author{Anoop Kunchukuttan, Pushpak Bhattacharyya \\
Center For Indian Language Technology \\
Department of Computer Science \& Engineering \\
Indian Institute of Technology Bombay \\
{\tt \{anoopk,pb\}@cse.iitb.ac.in}
}

\date{}

\begin{document}
\maketitle
\begin{abstract}
We explore the use of segments learnt using Byte Pair Encoding (referred to as \textit{BPE units}) as basic units for statistical machine translation between \textit{related} languages and compare it with \textit{orthographic syllables}, which are currently the best performing basic units for this translation task. BPE identifies the most frequent character sequences as basic units, while orthographic syllables are linguistically motivated pseudo-syllables. We show that BPE units modestly outperform orthographic syllables as units of translation, showing up to 11\% increase in BLEU score. While orthographic syllables can be used only for languages whose writing systems use vowel representations, BPE is writing system independent and we show that BPE outperforms other units for non-vowel writing systems too. Our results are supported by extensive experimentation spanning multiple language families and writing systems. 
\end{abstract}

\section{Introduction}
\label{sec:intro}

The term, \textit{related languages}, refers to languages that exhibit lexical and structural similarities on account of sharing a \textbf{common ancestry} or being in \textbf{contact for a long period of time} \cite{bhattacharyya2016statistical}. Examples of languages related by common ancestry are Slavic and Indo-Aryan languages. Prolonged contact leads to convergence of linguistic properties even if the languages are not related by ancestry and could lead to the formation of \textit{linguistic areas} \cite{thomason2000linguistic}. Examples of such linguistic areas are the Indian subcontinent \cite{emeneau1956india}, Balkan \cite{trubetzkoy1928proposition} and Standard Average European \cite{haspelmath2001european} linguistic areas. Genetic as well as contact relationship lead to related languages sharing vocabulary and structural features.

There is substantial government, commercial and cultural communication among people speaking related languages (Europe, India and South-East Asia being prominent examples and linguistic regions in Africa possibly in the future). As these regions  integrate more closely and move to a digital society, translation between \textit{related} languages is becoming an important requirement. In addition, translation to/from related languages to a \textit{lingua franca} like English is also very important. However, despite significant communication between people speaking related languages, most of these languages have few parallel corpora resources. It is therefore important to leverage the relatedness of these languages to build good-quality statistical machine translation (SMT) systems given the lack of parallel corpora. 

Modelling lexical similarity among related languages is the key to building good-quality SMT systems with limited parallel corpora. \textbf{Lexical  similarity} implies related languages share many words with similar form (spelling/pronunciation) and meaning \textit{e.g.} \texttt{blindness} is \texttt{andhapana} in Hindi, \texttt{aandhaLepaNaa} in Marathi. These words could be cognates, lateral borrowings or loan words from other languages. 

\textit{Subword level transformations} are an effective way for translation of such shared words. In this work, we propose use of Byte Pair Encoding (BPE) \cite{gage1994bpe,sennrich2016neural}, a encoding method inspired from text compression literature, to learn basic translation units for translation between related languages. In previous work, the basic units of translation are  either linguistically motivated (word, morpheme, syllable, etc.) or ad-hoc choices (character n-gram). In contrast, BPE is motivated by \textbf{statistical properties of text}.

The major contributions of our work are: 

\begin{itemize}[noitemsep]
\item We show that BPE units \textbf{modestly outperform orthographic syllable units} \cite{kunchukuttan2016orthographic}, the best performing basic unit for translation between related languages, resulting in up to 11\% improvement in BLEU score. 
\item Unlike orthographic syllables, BPE units are \textbf{writing system independent}. Orthographic syllables can only be applied to alphabetic and abugida writing systems. We show BPE units improve translation over word and morpheme level models for languages using \textit{abjad} and \textit{logographic} writing systems.  Average BLEU score improvements  of 18\% and 6\% over a baseline word-level model for language pairs involving abjad and logographic writing systems respectively were observed. 
\item Like orthographic syllables, BPE units outperform character, morph and word units when the language pairs show relatively less
lexical similarity or belong to different language families (but have sufficient contact relation).
\item While orthographic syllables approximate true syllables, we observe that BPE units learnt from the corpus \textbf{span various linguistic entities} (syllables, suffixes, morphemes, words, \etc). This may enable BPE level models to learn translation mappings at various levels simultaneously.
\item We have reported results over a large number of languages (16 language pairs and 17 languages) which span 4 major language families and 10 writing systems of various types. To the best of our knowledge, this is the largest experiment for translation over related languages and the \textbf{broad coverage strongly supports our results}. 
\item We also show BPE units outperform other translation units in a \textbf{cross-domain translation} task. 
\end{itemize}

The paper is organized as follows. Section \ref{sec:related} discusses related work. Section \ref{sec:bpe} discusses why BPE is a promising method for learning subword units and describes how we train BPE unit level translation models. Section \ref{sec:exp_setup} describes our experimental set-up. Section \ref{sec:results} reports the results of our experiments and analyses the results. Based on experimental results, we analyse why BPE units outperform other units in Section \ref{sec:why_bpe}. Section \ref{sec:conclusion} concludes the paper by summarizing our work and discussing further research directions.  

\section{Related Work}
\label{sec:related}

There are two broad set of approaches that have been explored in the literature for translation between related languages that leverage lexical similarity between source and target languages. 

The first approach involves \textbf{transliteration of source words} into the target languages. This can done by transliterating the untranslated words in a post-processing step \cite{nakov2012combining,kunchukuttan2014icon}, a technique generally used for handling named entities in SMT.  However, transliteration candidates cannot be scored and tuned along with other features used in the SMT system. This limitation can be overcome by integrating the transliteration module into the decoder \cite{durrani2010hindi}, so both translation and transliteration candidates can be evaluated and scored simultaneously. This also allows transliteration \vs translation choices to be made. 

Since a high degree of similarity exists at the subword level between related languages, the second approach looks at \textbf{translation with subword level basic units}. Character-level SMT has been explored for \text{very closely} related languages like \textit{Bulgarian-Macedonian, Indonesian-Malay, Spanish-Catalan} with modest success \cite{vilar2007can,tiedemann2009character,tiedemann2013analyzing}. Unigram-level learning provides very little context for learning translation models \cite{tiedemann2012character}. The use of character n-gram units to address this limitation leads to data sparsity for higher order n-grams and provides little benefit \cite{tiedemann2013analyzing}. These results were demonstrated primarily for very close European languages. \newcite{kunchukuttan2016orthographic} proposed \textit{orthographic syllables}, a linguistically-motivated variable-length unit, which approximates a syllable. This unit has outperformed character n-gram, word and morpheme level models as well as transliteration post-editing approaches mentioned earlier. They also showed orthographic syllables can outperform other units even when: (i) the lexical distance between related languages is reasonably large, (ii) the languages do not have a genetic relation, but only a contact relation.  

Recently, subword level models have also generated interest for neural machine translation (NMT) systems. The motivation is the need to limit the \textbf{vocabulary of neural MT systems} in encoder-decoder architectures \cite{sutskever2014sequence}. It is in this context that Byte Pair Encoding, a data compression method \cite{gage1994bpe}, was adapted to learn subword units for NMT \cite{sennrich2016neural}. Other subword units for NMT have also been proposed: character \cite{chung2016character}, Huffman encoding based units \cite{chitnis2015variable}, wordpieces \cite{schuster2012japanese,wu2016googlenmt}. Our hypothesis is that such subword units learnt from corpora are particularly suited for translation between related languages. In this paper, we test this hypothesis by using BPE to learn subword units. 

\section{BPE for related languages}
\label{sec:bpe}

We discuss why BPE is a promising method for learning subword units (subsections \ref{sec:motivation} and \ref{sec:compare_os}) and describe how we trained our BPE unit level translation models (subsections \ref{sec:bpe_algo} and \ref{sec:train_subword}). 

\subsection{Motivation}
\label{sec:motivation}

Byte Pair Encoding is a data compression algorithm which was first adapted for Neural Machine Translation by \newcite{sennrich2016neural}. For a given language, it is used to build a \textbf{vocabulary} relevant to translation by \textit{discovering the most frequent character sequences}  in the language. 

For NMT, BPE enables efficient, high quality, open vocabulary translation by (i) limiting core vocabulary size, (ii) representing the most frequent words as atomic BPE units and rare words as compositions of the atomic BPE units. These benefits of BPE are not particular to NMT, and apply to SMT between related languages too. Given the lexical similarity between related languages, we would like to \textit{identify a small, core vocabulary of subwords} from which words in the language can be composed. These subwords represent stable, frequent patterns (possibly linguistic units like syllables, morphemes, affixes) for which mappings exist in other related languages. This alleviates the need for word level translation. 

\subsection{Comparison with orthographic syllables}
\label{sec:compare_os}

We primarily compare BPE units with orthographic syllables (OS) \cite{kunchukuttan2016orthographic}, which are good translation units for related languages. The \textit{orthographic syllable} is a sequence of one or more consonants followed by a vowel, \ie a \textit{C$^+$V} unit, which approximates a linguistic syllable (\eg \textit{spacious} would be segmented as \textit{spa ciou s}). Orthographic syllabification is rule based and applies to writing systems which represent vowels (alphabets and abugidas). 

Both OS and BPE units are variable length units which provide longer and more relevant context for translation compared to character n-grams. In contrast to orthographic syllables, the BPE units are highly frequent character sequences reflecting the underlying statistical properties of the text. Some of the character sequences discovered by the BPE algorithm may be different linguistic units like syllables, morphemes and affixes. Moreover, BPE  can be applied to text in any writing system. 

\subsection{The BPE Algorithm}
\label{sec:bpe_algo}

We briefly summarize the BPE algorithm (described at length in \newcite{sennrich2016neural}). The input is a monolingual corpus for a language (one side of the parallel training data, in our case). We start with an \textit{initial vocabulary} \viz the characters in the text corpus. The vocabulary is updated using an iterative greedy algorithm. In every iteration, the most frequent bigram (based on current vocabulary) in the corpus is added to the vocabulary (the \textit{merge} operation). The corpus is again encoded  using the updated vocabulary and this process is repeated for a pre-determined number of merge operations. The number of merge operations is the only hyperparameter to the system which needs to be tuned. A new word can be segmented by looking up the learnt vocabulary. For instance, a new word \texttt{scion} may be segmented as \texttt{sc ion} after looking up the learnt vocabulary, assuming \texttt{sc} and \texttt{ion} as BPE units learnt during training. 

\subsection{Training subword level translation model}
\label{sec:train_subword}

We train subword level phrase-based SMT models between related languages. Along with BPE level, we also train PBSMT models at morpheme and OS levels for comparison. 

For BPE, we learn the vocabulary separately for the source and target languages using the respective part of the training corpus. We segment the data into subwords during pre-processing and indicate word boundaries by a boundary marker (\_) as shown in the example below. The boundary marker helps keep track of word boundaries, so the word level representation can be reconstructed after decoding.

{\small word}: {\scriptsize\texttt{Childhood means simplicity .}}  \\
{\small subword}: {\scriptsize\texttt{Chi ldhoo d \_ mea ns \_ si mpli ci ty \_ .}}

While building phrase-based SMT models at the subword level, we use (a) monotonic decoding since related languages have similar word order, (b) higher order languages models (10-gram) since data sparsity is a lesser concern owing to small vocabulary size \cite{vilar2007can}, and (c) word level tuning (by post-processing the decoder output during tuning) to optimize the correct translation metric \cite{nakov2012combining}. Following decoding, we used a simple method to regenerate words from subwords (desegmentation): concatenate subwords between consecutive occurrences of boundary marker characters.

\section{Experimental Setup}
\label{sec:exp_setup}

\begin{table}

\begin{subtable}{\columnwidth}
\setlength{\tabcolsep}{4pt}
\centering
{\small
\begin{tabular}{|ll|ll|ll|}
\hline
ben  &  Bengali  & kok  &  Konkani & pan  &  Punjabi \\ 
bul  &  Bulgarian & kor  &  Korean  & swe  &  Swedish \\ 
dan  &  Danish & mac  &  Macedonian & urd  &  Urdu  \\ 
hin  &  Hindi   & mar  &  Marathi  & tam  &  Tamil  \\ 
ind  &  Indonesian & mal  &  Malayalam  & tel  &  Telugu  \\ 
jpn  &  Japanese & may  &  Malay &  &  \\ 
\hline
\end{tabular}
}
\caption{List of languages used in experiments along with ISO 639-3 codes. These codes are used in the paper.}
\label{lbl:lang_list}
\end{subtable}

\vspace{0.3cm}

\begin{subtable}{\columnwidth}
\setlength\tabcolsep{3.5 pt}
\centering
\small{
\begin{tabular}{llll}
\toprule
\multicolumn{2}{c}{Language Family} & \multicolumn{2}{c}{Type of writing system }\\
\midrule
Dravidian & mal,tam,tel & \multirow{2}{*}{Alphabet} & dan$^1$,swe$^1$,may$^1$ \\
\multirow{2}{*}{Indo-Aryan} & hin,urd,ben &  & ind$^1$,buc$^2$,mac$^2$ \\
 & kok,mar,pan & \multirow{2}{*}{Abugida} & mal,tam,tel,hin   \\
Slavic & bul,mac & & ben,kok,mar,pan   \\
Germanic & dan,swe & Syllabic & kor  \\
Polynesian & may,ind & Logographic & jpn  \\
Altaic & jpn,kor & Abjad & urd \\
\bottomrule
\end{tabular}
}
\caption{Classification of the languages and writing systems. (i) Indo-Aryan, Slavic and Germanic belong to the larger Indo-European language family.  (ii) Alphabetic writing systems used by selected languages: Latin$^1$ and Cyrillic$^2$.}
\label{tbl:language_classify}
\end{subtable}
\caption{Languages under experiments: details}
\label{lbl:language_details}
\end{table}

We trained translation systems over the following basic units: character, morpheme, word, orthographic syllable and BPE unit. 
In this section, we summarize the languages and writing systems chosen for our experiments, the datasets used and the experimental configuration of our translation systems, and the evaluation methodology. 

\subsection{Languages and writing systems}

Our experiments spanned a diverse set of languages: 16 language pairs, 17 languages and 10 writing systems. Table \ref{lbl:language_details} summarizes the key aspects of the languages involved in the experiments. 

The chosen languages span 4 major language families (6 major sub-groups:  Indo-Aryan, Slavic and Germanic belong to the larger Indo-European language family). The languages exhibit diversity in word order and morphological complexity. Of course, between related languages, word order and morphological properties are similar. The classification of Japanese and Korean into the Altaic family is debated, but various lexical and grammatical similarities are indisputable, either due to genetic or cognate relationship \cite{martine2005isjapanese,vovin2010korea}. However, the source of lexical similarity is immaterial to the current work. For want of a better classification, we use the name \textit{Altaic} to indicate relatedness between Japanese and Korean. 

The chosen language pairs also exhibit varying levels of lexical similarity.  Table \ref{tbl:accuracy_bleu} shows an indication of the lexical similarity between them in terms of the Longest Common Subsequence Ratio (LCSR) \cite{melamed1995automatic}. The LCSR has been computed  over the parallel training sentences at character level (shown only for language pairs where the writing systems are the same or can be easily mapped in order to do the LCSR computation). At one end  of the spectrum, Malayalam-India, Urdu-Hindi, Macedonian-Bulgarian are dialects/registers of the same language and exhibit high lexical similarity. At the other end, pairs like Hindi-Malayalam belong to different language families, but show  many lexical and grammatical similarities due to contact for a long time \cite{subbarao2012south}.

The chosen languages cover 5 types of writing systems. Of these, alphabetic and abugida writing systems represent vowels, logographic writing systems do not have vowels. The use of vowels is optional in abjad writing systems and depends on various factors and conventions.  For instance, Urdu word segmentation can be very inconsistent \cite{durrani2010urdu} and generally short vowels are not denoted. The Korean \textit{Hangul} writing system is syllabic, so the vowels are implicitly represented in the characters. 

\subsection{Datasets}

Table \ref{tbl:parallel_corpus_details} shows train, test and tune splits of the parallel corpora used. The Indo-Aryan and Dravidian language parallel corpora are obtained from the multilingual Indian Language Corpora Initiative (ILCI) corpus \cite{jha2012ilci}. Parallel corpora for other pairs were obtained from the \textit{OpenSubtitles2016} section of the OPUS corpus collection \cite{tiedemann2009news}. Language models for word-level systems were trained on the target side of training corpora  plus additional monolingual corpora from various sources (See Table \ref{tbl:monolingual_corpus_details} for details). We used just the target language side of the parallel corpora for character, morpheme, OS and BPE-unit level LMs. 

\begin{table}[t]
\begin{subtable}{\columnwidth}
\centering
{\small
\begin{tabular}{lrrr}
\toprule
\hh{Language Pair} & \hh{train} & \hh{tune} & \hh{test} \\
\midrule 
ben-hin,pan-hin, & \multirow{3}{*}{44,777} & \multirow{3}{*}{1000} & \multirow{3}{*}{2000} \\
kok-mar, & & & \\
mal-tam,tel-mal, & & & \\
hin-mal,mal-hin & & & \\
\midrule
urd-hin,ben-urd & \multirow{2}{*}{38,162} & \multirow{2}{*}{843} & \multirow{2}{*}{1707} \\
urd-mal,mal-urd & & & \\
\midrule
bul-mac & 150k & 1000 & 2000 \\
dan-swe & 150k & 1000 & 2000 \\
may-ind & 137k & 1000 & 2000 \\
\midrule
kor-jpn,jpn-kor & 69,809 & 1000 & 2000 \\
\bottomrule
\end{tabular}
}
\caption{Parallel Corpora Size (no. of sentences)}
\label{tbl:parallel_corpus_details}
\end{subtable}

\vspace{0.3cm}

\begin{subtable}{\columnwidth}
\setlength\tabcolsep{3.5 pt}
\centering
{\scriptsize
\begin{tabular}{lrlr}
\toprule
\hh{Language} & \hh{Size} & \hh{Language} & \hh{Size} \\
\midrule 
hin \cite{hindencorp:lrec:2014} & 10M & urd \cite{jawaid2014urdu} & 5M \\
tam \cite{ramasamy-bojar-zabokrtsky:2012:MTPIL} & 1M & mar (news websites) & 1.8M \\
mal \cite{quasthoff2006corpus} & 200K & swe (OpenSubtitles2016) & 2.4M \\
mac \cite{tiedemann2009news} & 680K & ind \cite{tiedemann2009news} & 640K \\
\bottomrule
\end{tabular}
}
\caption{Details of additional monolingual corpora for training word-level language models (source and size in number of sentences)}
\label{tbl:monolingual_corpus_details}
\end{subtable}
\caption{Training Corpus Statistics}
\label{tbl:corpus_details}
\end{table}

\subsection{System details} 
We trained phrase-based SMT systems using the \textit{Moses} system \cite{koehn2007moses}, with the \textit{grow-diag-final-and} heuristic for extracting phrases, and Batch MIRA \cite{cherry2012batch} for tuning (default parameters). We trained 5-gram LMs with Kneser-Ney smoothing for word and morpheme level models and 10-gram LMs for character, OS and BPE-unit level models. Subword level representation of sentences is long, hence we speed up decoding by using cube pruning with a smaller beam size (pop-limit=1000). This setting has been shown to have minimal impact on translation quality \cite{kunchukuttan2016faster}.

We used unsupervised morphological-segmenters for generating morpheme representations (trained using \textit{Morfessor} \cite{smit2014morfessor}). For Indian languages, we used the models distributed as part of the \textit{Indic NLP Library}\footnote{http://anoopkunchukuttan.github.io/indic\_nlp\_library} \cite{kunchukuttan2014icon}. We used orthographic syllabification rules from the \textit{Indic NLP Library} for Indian languages, and custom rules for Latin and Slavic scripts. For training BPE models, we used the \textit{subword-nmt}\footnote{https://github.com/rsennrich/subword-nmt} library. We used \textit{Juman}\footnote{http://nlp.ist.i.kyoto-u.ac.jp/EN/index.php?JUMAN} and \textit{Mecab}\footnote{https://bitbucket.org/eunjeon/mecab-ko} for Japanese and Korean tokenization respectively. 

For mapping characters across Indic scripts, we used the method described by \newcite{kunchukuttan2015brahmi} and implemented in the \textit{Indic NLP Library}.

\subsection{Evaluation}
The primary evaluation metric is \textit{word-level} BLEU \cite{papineni2002bleu}. We also report LeBLEU \cite{virpioja2015lebleu} scores as an alternative evaluation metric. LeBLEU is a variant of BLEU that does an edit-distance based, soft-matching of words and has been shown to be better for morphologically rich languages. We used bootstrap resampling for testing statistical significance \cite{koehn2004statistical}.

\section{Results and Analysis}
\label{sec:results}

This section describes the results of various experiments and analyses them. A comparison of BPE with other units across languages and writing systems, choice of number of merge operations and effect of domain change and training data size are studied. We also report initial results with a joint bilingual BPE model. 

\subsection{Comparison of BPE with other units}

\begin{table*}[t]
\centering
{\small
\begin{tabular}{lrrrrrrrrrrr}
\toprule
\multicolumn{2}{c}{\textbf{Language Pair}} &  \multicolumn{5}{c}{\textbf{BLEU}} &  \multicolumn{5}{c}{\textbf{LeBLEU}} \\
\cmidrule(lr){1-2}
\cmidrule(lr){3-7}
\cmidrule(lr){8-12}
\thead{Src-Tgt} & \thead{LCSR} & \thead{C} & \thead{W} & \thead{M} & \thead{O} & \thead{B$_{match}$}  & \thead{C} & \thead{W} & \thead{M} & \thead{O} & \thead{B$_{match}$} \\
\cmidrule(lr){1-2}
\cmidrule(lr){3-7}
\cmidrule(lr){8-12}
ben-hin & 52.30 & 27.95 & 32.47          & 32.17 & \textbf{33.54}               & 33.22           & 0.672 & 0.682          & 0.708 & 0.715          & \textbf{0.716} \\
pan-hin & 67.99 & 71.26 & 70.07          & 71.29 & \textbf{72.41}               & 72.22           & 0.905 & 0.871          & 0.899 & 0.906          & \textbf{0.907} \\
kok-mar & 54.51 & 19.83 & 21.30          & 22.81 & 23.43                        & \textbf{23.63}  & 0.632 & 0.636          & 0.659 & \textbf{0.671} & 0.665          \\
mal-tam & 39.04 & 4.50  & 6.38           & 7.61  & 7.84                         & \textbf{8.67}$\dagger$   & 0.311 & 0.314          & 0.409 & 0.447          & \textbf{0.465} \\
tel-mal & 39.18 & 6.00  & 6.78           & 7.86  & 8.50                         & \textbf{8.79}   & 0.346 & 0.314          & 0.383 & 0.439          & \textbf{0.443} \\
hin-mal & 33.24 & 6.28  & 8.55           & 9.23  & 10.46                        & \textbf{10.73}  & 0.324 & 0.393          & 0.436 & \textbf{0.477} & 0.468          \\
mal-hin & 33.24 & 12.33 & 15.18          & 17.08 & 18.44                        & \textbf{20.54}  & 0.444 & 0.460          & 0.528 & 0.551          & \textbf{0.565} \\
\cmidrule(lr){1-2}
\cmidrule(lr){3-7}
\cmidrule(lr){8-12}
urd-hin & -     & 52.57 & 55.12          & 52.87 & NA                           & \textbf{55.55}  & 0.804 & 0.795          & 0.792 & NA             & \textbf{0.823} \\
ben-urd & -     & 18.16 & 27.06          & 27.31 & NA                           & \textbf{28.06}  & 0.607 & 0.660          & 0.671 & NA             & \textbf{0.692} \\
urd-mal & -     & 3.13  & 6.49           & 7.05  & NA                           & \textbf{8.44}   & 0.247 & 0.350          & 0.379 & NA             & \textbf{0.416} \\
mal-urd & -     & 8.90  & 13.22          & 15.30 & NA                           & \textbf{18.48}  & 0.444 & 0.454          & 0.522 & NA             & \textbf{0.568} \\
\cmidrule(lr){1-2}
\cmidrule(lr){3-7}
\cmidrule(lr){8-12}
bul-mac & 62.85 & 20.61 & 21.20          & -     & \textbf{21.95}               & 21.73           &  0.603 & 0.606          &  -     & \textbf{0.613} & 0.599          \\
dan-swe & 63.39 & 35.36 & 35.13          & -     & 35.46                   & \textbf{35.77}  & 0.692 & \textbf{0.694} &  -     & 0.682          & 0.682          \\
may-ind & 73.54 & 60.50 & \textbf{61.33} & -     & 60.79                        & 59.54$\dagger$           & 0.827 & \textbf{0.832} &  -     & 0.828          & 0.825   \\      
\cmidrule(lr){1-2}
\cmidrule(lr){3-7}
\cmidrule(lr){8-12}
kor-jpn & -     & 8.51     & 9.90           & -     & NA                           & \textbf{10.23}  & 0.396     & 0.372          &  -     & NA             & \textbf{0.408} \\
jpn-kor & -     & 8.17     & 8.44           & -     & NA                           & \textbf{9.02}   & 0.372     & 0.350          &  -     & NA             & \textbf{0.374} \\
\bottomrule
\end{tabular}
}
\caption{Translation accuracies for various translation units (BLEU and LeBLEU scores reported). The reported scores are:- \textbf{W}: word-level, \textbf{M}: morpheme, \textbf{O}: orthographic syllable, \textbf{B$_{match}$}: BPE units with number of merge operations selected to match vocabulary size of OS encoding. See discussion related to exceptions for pairs involving Urdu, Korean and Japanese. (a) The values marked in \textbf{bold} indicate best score for a language pair (b) \textbf{LCSR} indicates lexical similarity  (c) \textit{NA: Not Applicable.} (d) $\dagger$ indicates that difference in BLEU scoresbetween \hl{B$_{match}$} and \hl{O} are statistically significant ($p<0.05$)}
\label{tbl:accuracy_bleu}
\end{table*}

Table \ref{tbl:accuracy_bleu} shows translation accuracies of all the language pairs under experimentation for different translation units, in terms of BLEU as well as LeBLEU scores. The number of BPE merge operations was chosen such that the resultant vocabulary size would be equivalent to the vocabulary size of the orthographic syllable encoded corpus. Since we could not do orthographic syllabification for Urdu, Korean and Japanese, we selected the merge operations as follows: For Urdu, number of merge operations were selected based on Hindi OS vocabulary since Hindi and Urdu are registers of the same language. 
For Korean and Japanese, the number of BPE merge operations was set to 3000, discovered by tuning on a separate validation set.

Our major observations are described below (based on BLEU scores):

\noindent $\bullet\,$ BPE units are clearly better than the traditional word and morpheme representations. The average BLEU score improvement is 15\% over word-based results and 11\% over morpheme-based results. The only exception is Malay-Indonesian, which are registers of the same language.

\noindent $\bullet\,$ BPE units also  show modest improvement over the recently proposed orthographic syllables over most language pairs (average improvement of 2.6\% and maximum improvement of up to 11\%). The improvements are not statistically significant  for most language pairs. The only exceptions are Bengali-Hindi, Punjabi-Hindi and Malay-Indonesian - all these languages pairs have relatively less morphological affixing (\textit{Bengali-Hindi, Punjabi-Hindi}) or are registers of the same language (Malay-Indonesian). For Bengali-Hindi and Punjabi-Hindi, the BPE unit translation accuracies are quite close to OS level accuracies.  Since OS level models have been shown to be better than character level models \cite{kunchukuttan2016orthographic}, BPE units are better than character level models by transitivity.

\noindent $\bullet\,$ BPE units also outperform other units for translation between language pairs belonging to different language pairs, but having a long contact relationship \viz Malayalam-Hindi and Hindi-Malayalam.

\noindent $\bullet\,$ It is worth mentioning that BPE units provide a substantial benefit over OS units when translation involves a morphologically rich language. In translations involving Malayalam, Tamil and Telugu, average accuracy improvement of 6.25\% were observed.

The LeBLEU scores also show the same trends as the BLEU scores. 

\subsection{Applicability to different writing systems}
The utility of orthographic syllables as translation units is limited to languages that use writing systems which represent vowels. Alphabetic and abugida writing systems fall into this category. On the other hand, logographic writing systems (Japanese Kanji, Chinese) and abjad writing systems (Arabic, Hebrew, Syriac, etc.) do not represent vowels. To be more precise, abjad writing systems may represent some/all vowels depending on  language, pragmatics and conventions. Syllabic writing systems like Korean Hangul do not explicitly represent vowels, since the basic unit (the syllable) implicitly represents the vowels. The major advantage of Byte Pair Encoding is its \textbf{writing system independence} and our results show that BPE encoded units are useful for translation involving abjad (Urdu uses an extended Arabic writing system), logographic (Japanese Kanji) and syllabic (Korean Hangul) writing systems. For language pairs involving Urdu, there is an 18\% average improvement over word-level and 12\% average improvement over morpheme-level translation accuracy. For Japanese-Korean language pairs, an average improvement of 6\% in translation accuracy  over a word-level baseline is observed. 

\subsection{Choosing number of BPE merges}

\begin{table}[t]
\setlength\tabcolsep{3.5 pt}
\centering
{\small
\begin{tabular}{lrrrrrr}
\toprule
& \thead{O} & \thead{B$_{match}$} & \thead{B$_{1k}$} & \thead{B$_{2k}$} & \thead{B$_{3k}$} & \thead{B$_{4k}$} \\
\midrule
ben-hin & \hl{33.54} & 33.22 & 33.16 & 33.25 & \hll{33.30} & 32.99 \\ 
pan-hin & \hl{72.41} & 72.22 & \hll{72.28} & 72.19 & 72.08 & 71.94 \\ 
kok-mar & 23.43 & 23.63 & \hl{23.84} & 23.73 & 23.79 & 23.30 \\ 
mal-tam & 7.84 & 8.67 & 8.66 & 8.71 & 8.63 & \hl{8.74} \\ 
tel-mal & 8.50 & 8.79 & 8.99 & 8.83 & \hl{9.12} & 8.76 \\ 
hin-mal & 10.46 & 10.73 & \hl{10.96} & 10.89 & 10.61 & 10.55 \\ 
mal-hin & 18.44 & 20.54 & \hl{21.23} & 20.53 & 20.64 & 20.19 \\ 
\midrule
urd-hin & NA & 55.55 & \hl{55.69} &  55.49 & 55.57 & 55.47 \\
ben-urd & NA & 28.06 & 28.12 &  \hl{28.19} & 28.03 & 27.93 \\ 
urd-mal & NA & 8.44 & 8.22 &   8.04 & 8.02 & \hl{8.57} \\ 
mal-urd & NA & 18.48 & 18.72 &  18.47 & \hl{18.79} & 18.18 \\  
\midrule
bul-mac & 21.95 & 21.73 & 21.74 & \hl{22.27} & 21.95 & 21.94 \\ 
dan-swe & 35.46 & 35.77 & 36.38 & 36.18 & \hl{36.61} & 36.2 \\ 
may-ind & \hl{60.79} & 59.54 & \hll{60.63} & 60.24 & 60.35 & 60.15 \\ 
\midrule
kor-jpn & NA & NA &  10.13 & 9.8 & \hl{10.23} & 9.92 \\ 
jpn-kor & NA & NA & \hl{9.29} & 9.23 & 9.02 & 8.96 \\ 
\bottomrule
\end{tabular}
}
\caption{Translation accuracies for BPE models trained with different number of merge operations (BLEU). Underlined scores indicate the best BPE configuration when OS is the best-performing for a language pair.}
\label{tbl:bpe_mergesize_bleu}
\end{table}

The above mentioned results for BPE units do not explore optimal values of the number of merge operations. This is the only hyper-parameter that has to be selected for BPE. We experimented with number of merge operations ranging from 1000 to 4000 and the translation results for these are shown in Table \ref{tbl:bpe_mergesize_bleu}. Selecting the optimal value of merge operations lead to a modest, average increase of 1.6\% and maximum increase of 3.5\% in the translation accuracy over B$_{match}$ across different language pairs .

We also experimented with higher number of merge operations for some language pairs, but there seemed to be no benefit with a higher number of merge operations. Compared to the number of merge operations  reported by \newcite{sennrich2016neural} in a more general setting for NMT (60k), the number of merge operations is far less  for translation between related languages with limited parallel corpora. We must bear in mind that their goal was different: available parallel corpus was not an issue, but they wanted to handle as large a vocabulary as possible for open-vocabulary NMT. Yet, the low number of merge operations suggest that BPE encoding captures the core vocabulary required for translation between related tasks.

\subsection{Robustness to Domain Change} 

\begin{table}

\begin{subtable}[l]{\columnwidth}
\setlength\tabcolsep{3.5 pt}
\begin{tabular}{lrrrrr}
\toprule
\thead{Pair} & \thead{C} & \thead{W} & \thead{M} & \thead{O} &\thead{B$_{match}$} \\
\midrule
pan-hin & 58.07 & 58.95 & \textbf{59.71} & 57.95          & 59.66$^\dagger$          \\
kok-mar & 17.97 & 18.83 & 18.53          & \textbf{19.12} & 18.42$^\dagger$          \\
mal-tam & 4.12 & 5.49  & 5.84           & 5.93           & \textbf{6.75}$^\dagger$  \\
tel-mal & 3.11 & 3.26  & \textbf{4.06}  & 3.83           & 3.75           \\
hin-mal & 3.85 & 5.18  & 5.99           & 6.24           & \textbf{6.37}  \\
mal-hin & 8.42 & 9.92  & 11.12          & 13.36          & \textbf{14.45}$^\dagger$ \\
\bottomrule
\end{tabular}
\caption{BLEU scores}
\label{tbl:agr_accuracy_bleu}
\end{subtable}

\begin{subtable}[l]{\columnwidth}
\setlength\tabcolsep{3.5 pt}
\begin{tabular}{lrrrrr}
\toprule
\thead{Pair} & \thead{C} & \thead{W} & \thead{M} & \thead{O} &\thead{B$_{match}$} \\
\midrule
pan-hin & 0.869 & 0.825 & 0.868 & 0.863          & \textbf{0.876} \\
kok-mar & 0.647 & 0.641 & 0.643 & \textbf{0.665} & 0.653          \\
mal-tal & 0.301 & 0.261 & 0.378 & 0.452          & \textbf{0.475} \\
tel-mal & 0.246 & 0.198 & 0.238 & 0.297          & \textbf{0.300} \\
hin-mal & 0.281 & 0.336 & 0.354 & \textbf{0.404} & 0.384          \\
mal-hin & 0.439 & 0.371 & 0.466 & 0.548          & \textbf{0.565} \\
\bottomrule
\end{tabular}
\caption{LeBLEU scores}
\label{tbl:agr_accuracy_lebleu}
\end{subtable}

\caption{Translation accuracies for Agriculture Domain $^\dagger$ indicates statistically significant difference in BLEU score between \hl{O} and \hl{B$_{match}$}. BLEU score differences between \hl{B$_{match}$} and \hl{W} are also statistically significant (except Konkani-Marathi) ($p<0.05$)}
\label{tbl:agr_accuracy}

\end{table}

\begin{figure}[t]
\centering
\includegraphics[scale=0.5]{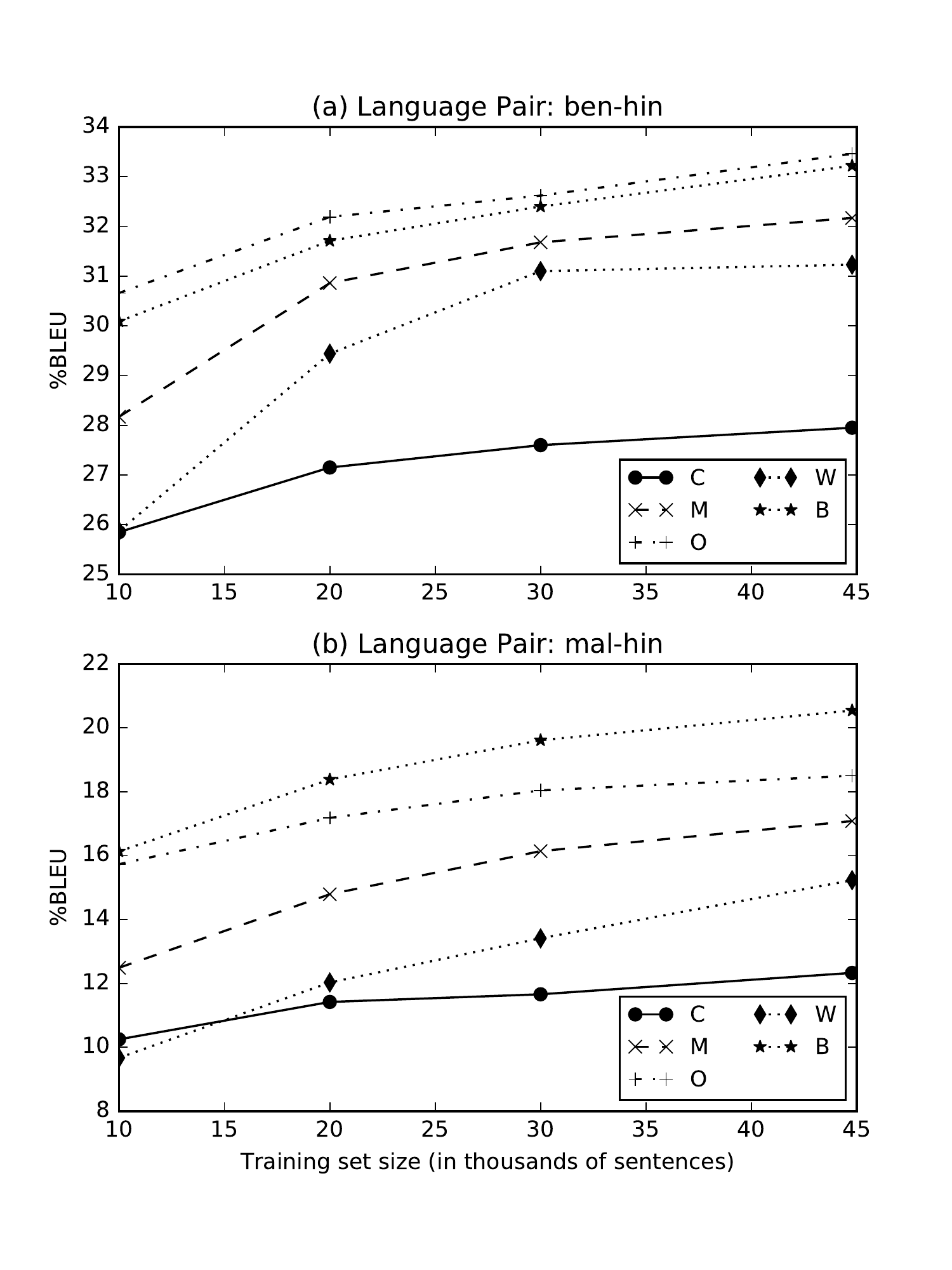}
\caption{Effect of training data size on translation accuracy for different basic units}
\label{fig:size_effect}
\end{figure}

Since we are concerned with low resource scenarios, a desirable property of subword units is robustness of the translation models to change of translation domain. \newcite{kunchukuttan2016orthographic} have shown that OS level models are robust to domain change. Since BPE units are learnt from a specific corpus, it is not guaranteed that they would also be robust to domain changes. To study the behaviour of BPE unit trained models, we also tested the translation models trained on tourism \& health domains on an agriculture domain test set of 1000 sentences (see Table \ref{tbl:agr_accuracy} for results). \textit{In this cross-domain translation scenario, the BPE level model outperforms the OS-level and word-level models for most language pairs}. The Konkani-Marathi pair alone shows a degradation using the OS level model. The BPE model is almost  on par with the OS level model for Telugu-Malayalam and Hindi-Malayalam.

\subsection{Effect of training data size} 
For different training set sizes, we trained SMT systems with various representation units (Figure \ref{fig:size_effect} shows the learning curves for two language pairs). BPE level models are better than OS, morpheme and word level across a range of dataset sizes. Especially when the training data is very small, the OS and BPE level models perform significantly better than the word and morpheme level models. For Malayalam-Hindi, the BPE level model is better than the OS level model at utilizing more training data. 

\subsection{Joint bilingual learning of BPE units}

\begin{table}[t]
\setlength\tabcolsep{3.5 pt}
\centering
{\small
\begin{tabular}{lrrrrr}
\toprule
& \thead{Best$_{prev}$} & \thead{JB$_{1k}$} & \thead{JB$_{2k}$} & \thead{JB$_{3k}$} & \thead{JB$_{4k}$} \\
\midrule
ben-hin & O  (33.46) & \hl{33.54} & 33.23 & 33.54 & 33.35 \\ 
pan-hin & O  (\hl{72.51}) & \hll{72.41} & 72.35 & 72.13 & 72.04 \\ 
kok-mar & B$_{1k}$ (23.84) & \hl{24.01} & 23.76 & 23.8 & 23.86 \\ 
mal-tam & B$_{4k}$ (8.74)  & 8.6 & \hl{8.82} & 8.74 & 8.72 \\ 
tel-mal & B$_{3k}$ (\hl{9.12})  & 8.47 & 8.84 & 8.89 & \hll{8.92} \\ 
hin-mal & B$_{1k}$ (10.96) & \hl{11.19} & 11.09 & 11.1 & 10.96 \\ 
mal-hin & B$_{1k}$ (\hl{21.23}) & 20.79 & \hll{21.22} & 21.12 & 21.06 \\ 
\midrule
bul-mac & B$_{2k}$ (\hl{22.27}) & 22.11 & 22.17 & 21.58 & \hll{22.24} \\ 
dan-swe & B$_{3k}$ (36.61) & 36.15 & \hl{36.86} & 36.51 & 36.71 \\
may-ind & O (61.24) & \hl{61.26} & 60.98 & 61.11 & 60.66 \\ 
\bottomrule
\end{tabular}
}
\caption{Translation accuracies for Joint BPE models trained with different number of merge operations (BLEU). The \textbf{Best$_{prev}$} indicates the best performing units and their accuracy scores from Tables \ref{tbl:accuracy_bleu} and \ref{tbl:bpe_mergesize_bleu} shown for comparison.}
\label{tbl:jointbpe_mergesize_bleu}
\end{table}

In the experiments discussed so far, we learnt the BPE vocabulary separately for the source and target languages. In this section, we describe our experiments with jointly learning BPE vocabulary over source and target language corpora as suggested by \newcite{sennrich2016neural}. The idea is to learn an encoding that is consistent across source and target languages and therefore helps alignment.  We expect a significant number of common BPE units between related languages. If source and target languages use the same writing system, then a joint model is created by learning BPE over concatenated source and target language corpus. If the writing systems are different, then we transliterate one corpus to another by one-one character mappings. This is possible between Indic scripts. But this scheme cannot be applied between Urdu and Indic scripts as well as between Korean Hangul and Japanese Kanji scripts.

Table \ref{tbl:jointbpe_mergesize_bleu} shows the results of the joint BPE model for language pairs where such a model is built. We do not see any major improvement over the monolingual BPE model due to the joint BPE model.

\section{Why are BPE units better than others?}
\label{sec:why_bpe}

\begin{table}[t]
\setlength\tabcolsep{3.5 pt}
\centering
\begin{tabular}{lrrrrr}
\toprule
\thead{Src-Tgt} & \thead{Word} & \thead{Morph} & \thead{BPE} & \thead{OS}  & \thead{Char}\\
\midrule
ben-hin           & 0.40           & 0.58            & 0.60                   & 0.62         & 0.71           \\
pan-hin           & 0.50           & 0.64            & 0.69                   & 0.70         & 0.72           \\
kok-mar           & 0.66           & 0.63            & 0.64                   & 0.67         & 0.74           \\
mal-tam           & 0.46           & 0.56            & 0.70                   & 0.71         & 0.77           \\
tel-mal           & 0.45           & 0.52            & 0.62                   & 0.64         & 0.78           \\
hin-mal           & 0.39           & 0.46            & 0.52                   & 0.58         & 0.79           \\
mal-hin           & 0.37           & 0.45            & 0.54                   & 0.60         & 0.71           \\
\bottomrule
\end{tabular}
\caption{Pearson's correlation coefficient between lexical similarity and translation accuracy (both in terms of LCSR at character level). \textit{This was computed over the test set between: (i) sentence level lexical similarity between source  and target sentences and (ii) sentence level translation accuracy between hypothesis and reference.}}
\label{tbl:corr_sim_acc}
\end{table}

The improved performance of BPE units compared to word-level and morpheme-level representations is easy to explain: with a limited vocabulary they \textbf{address the problem of data sparsity}. But character level models also have a limited vocabulary, yet they do not improve translation performance except for very close languages. Character level models learn character mappings effectively, which is sufficient for translating related languages which are very close to each other (translation is akin to transliteration in these cases). But they are not sufficient for translating related languages that are more divergent. In this case, translating cognates, morphological affixes, non-cognates \etc require a larger context. So, BPE and OS units --- which \textbf{provide more context} --- outperform character units.

A study of the correlation between lexical similarity and translation quality makes this evident (See Table \ref{tbl:corr_sim_acc}). We see that character models work best when the source and target sentences are lexically very similar.  The additional context decouples OS and BPE units from lexical similarity. Words and morphemes show the least correlation since they do not depend on lexical similarity.

\begin{table}
\centering
{\scriptsize
\begin{tabular}{llll}
\toprule
 & \thead{hin} & \thead{mar} & \thead{mal} \\ 
 \bottomrule 
\textbf{OS} & tI, stha & mA, nA & kka, nI \\ 
\textbf{Suffix} & ke, me.m  & ChyA, madhIla   & unnu, .e$\sim$Nkill.m \\ 
\textbf{Word} & paryaTaka, athavA & prAchIna, aneka & bhakShaN.m, yAtra \\ 
\bottomrule
\end{tabular}
}
\caption{Examples of BPE units for Indian languages. (ITRANS transliteration shown)}
\label{tbl:bpe_eg}
\end{table}

Why does BPE performs better than OS which also provides a larger contextual window for translation? While orthographic syllables represent just approximate syllables, we observe that BPE units also \textbf{represent higher level semantic units like frequent morphemes, suffixes and entire words}. Table \ref{tbl:bpe_eg} shows a few examples for some Indian languages. So, BPE level models can learn semantically similar translation mappings in addition to lexically similar mappings. In this way, BPE units enable the translation models to \textbf{balance the use of lexical similarity with semantic similarity}. This further decouples the translation quality from lexical similarity as seen from the table. BPE units also have an \textbf{additional degree of freedom} (choice of vocabulary size), which allows tuning for best translation performance. This could be important when larger parallel corpora are available, allowing larger vocabulary sizes.

\section{Conclusion \& Future Work}
\label{sec:conclusion}

We show that translation units learnt using BPE can outperform all previously proposed translation units, including the best-performing orthographic syllables, for SMT between related languages when limited parallel corpus is available. Moreover, BPE encoding is writing system independent, hence it can be applied to any language. Experimentation on a large number of language pairs spanning diverse language families and writing systems lend strong support to our results. We also show that BPE units are more robust to change in translation domain. They perform better for morphologically rich languages and extremely data scarce scenarios.

BPE seems to be beneficial because it enables discovery of translation mappings at various levels simultaneously (syllables, suffixes, morphemes, words, \etc). We would like to further pursue this line of work and investigate better translation units. This is also a question relevant to translation with subwords in NMT. NMT between related languages using BPE and similar encodings is also an obvious direction to explore.

Given the improved performance of the BPE-unit, tasks involving related languages \textit{viz.} pivot based MT, domain adaptation \cite{tiedemann2012character} and translation between a \textit{lingua franca} and related languages \cite{wang2012source} can be revisited with BPE units. 

\section*{Acknowledgments}
We thank the Technology Development for Indian Languages (TDIL) Programme and the Department of Electronics \& Information Technology, Govt. of India for their support. We also thank the  reviewers for their feedback. 

\bibliography{wmt2017_bpe}
\bibliographystyle{emnlp_natbib}

\end{document}